\documentclass{IEEEcsmag}

\usepackage[colorlinks,urlcolor=blue,linkcolor=blue,citecolor=blue]{hyperref}
\expandafter\def\expandafter\UrlBreaks\expandafter{\UrlBreaks\do\/\do\*\do\-\do\~\do\'\do\"\do\-}
\usepackage{upmath,color}

\usepackage{times}
\usepackage[numbers,sort&compress]{natbib}

\usepackage{graphicx}

\let\labelindent\relax
\usepackage{enumitem}
\usepackage{xcolor}

\usepackage{xspace}
\usepackage{soul}

\usepackage{circledsteps}
\pgfkeys{/csteps/inner color=white}
\pgfkeys{/csteps/fill color=black}

\newcommand{\eg}{{e.g.,}\xspace}
\newcommand{\viz}{{viz.,}\xspace}

\newcommand{\ci}{{\it (i) }}
\newcommand{\cii}{{\it (ii) }}
\newcommand{\ciii}{{\it (iii) }}

\newcommand{\ca}{{\it (a) }}
\newcommand{\cb}{{\it (b) }}
\newcommand{\cc}{{\it (c) }}

\definecolor{notecolor}{rgb}{0.8,0,0} 

\newcommand{\acmbfpar}[1]{{\vspace*{0.3\baselineskip}\noindent\bfseries #1\quad}}

\jvol{XX}
\jnum{XX}
\paper{8}

\begin{document}


\title{LLMs as On-demand Customizable Service}

\author{Souvika Sarkar}
\affil{Auburn University, Auburn,  AL, USA}

\author{Mohammad Fakhruddin Babar}
\affil{Washington State University, Pullman, WA, USA}

\author{Monowar Hasan}
\affil{Washington State University, Pullman, WA, USA}

\author{Shubhra Kanti Karmaker (Santu)}
\affil{Auburn University, Auburn,  AL, USA}

\markboth{LLMs as On-demand Customizable Service}{LLMs as On-demand Customizable Service}

\begin{abstract}
Large Language Models (LLMs) have demonstrated remarkable language understanding and generation capabilities. However, training, deploying, and accessing these models pose notable challenges, including resource-intensive demands, extended training durations, and scalability issues. To address these issues, we introduce a concept of \textit{hierarchical, distributed LLM architecture} that aims at enhancing the accessibility and deployability of LLMs across heterogeneous computing platforms, including general-purpose computers (\eg laptops) and IoT-style devices (\eg embedded systems). By introducing a ``layered'' approach, the proposed architecture enables on-demand accessibility to LLMs as a customizable service. This approach also ensures optimal trade-offs between the available computational resources and the user's application needs. We envision that the concept of hierarchical LLM will empower extensive, crowd-sourced user bases to harness the capabilities of LLMs, thereby fostering advancements in AI technology in general.

\end{abstract}

\maketitle


\chapteri{L}arge Language Models (LLMs), such as GPT-3.5~\cite{chatgpt}, LLaMA~\cite{touvron2023llama}, BLOOM~\cite{bloom}, have unleashed unparalleled language processing capabilities, serving as catalysts for developing various AI tools and enabling seamless knowledge transfer across diverse NLP tasks, including automatic news summarization~\cite{zhang2023benchmarking}, efficient healthcare research~\cite{sallam2023chatgpt}, software bug solving~\cite{surameery2023use}, zero-shot classification~\cite{sarkar2022exploring, sarkar2023zero} and machine translation~\cite{jiao2023chatgpt}. 
While LLMs offer remarkable advantages, their widespread adoption encounters substantial challenges, primarily stemming from their enormous size. These models boast an astronomical number of parameters, necessitating considerable computational resources. Unfortunately, such resources are often lacking on local devices, rendering LLMs less accessible and casting doubts on their customizability for specific applications, particularly on devices with limited computational capabilities. The intricacies of training, deploying, and consistently updating LLMs further compound these challenges. 
In response to the challenges presented by LLMs, we present an innovative idea: a \textbf{hierarchical, distributed architecture} (Sec.~\ref{sec:architecture}). Our proposed solution aimed to improve the accessibility and utility of LLMs in various settings through the following aspects.


    \acmbfpar{Hierarchical Organization of Knowledge.} By hierarchically structuring an LLM, vast knowledge learned from big data corpora is distributed across multiple layers based on target language (\eg English, Spanish), target application domains (\eg Medical, Sports), and application-oriented sub-domains (\eg Soccer, Baseball). The basic idea is that each node in this hierarchy represents a language model, where nodes in the upper layer of the hierarchy represent models that are general-purpose (and larger), while nodes at the lower layers of the hierarchy represent models that are small and domain-specific. This organization allows for a more efficient organization of information, reducing redundancy and eliminating the need for every application to store the entire model. As a result, the size of each application-specific language model is more manageable.
    
   \acmbfpar{Enhanced Customization.} The hierarchical architecture permits users to custom-select LLMs according to their specific requirements. Rather than deploying a monolithic, one-size-fits-all model with an enormous parameter count, users can select an LLM from the ``right/optimal'' layer and further configure it to suit their application.
    
    \acmbfpar{Efficient Resource Management.} The architecture optimizes resource (computing power, memory, battery capacity) allocation by allowing users to choose a language model that matches their hardware capabilities. Local devices with limited computational resources can opt for a smaller, more resource-friendly language model from lower layers of the hierarchy. This prevents over-commitment of resources and ensures that LLMs can run effectively on various devices, from personal computers to low-power embedded devices.
    
    \acmbfpar{Scalability.} The hierarchical structure also supports scalability. As the demands of an application grow in terms of its knowledge base, users can upgrade their application-specific-language model by selecting a model from a higher layer that offers \textit{enhanced capabilities} and covers a larger knowledge base. This scalability ensures that applications can handle more complex tasks without transitioning to an entirely new model architecture.

\begin{figure*}[!htb]
\centering
        \includegraphics[width=0.8\linewidth]{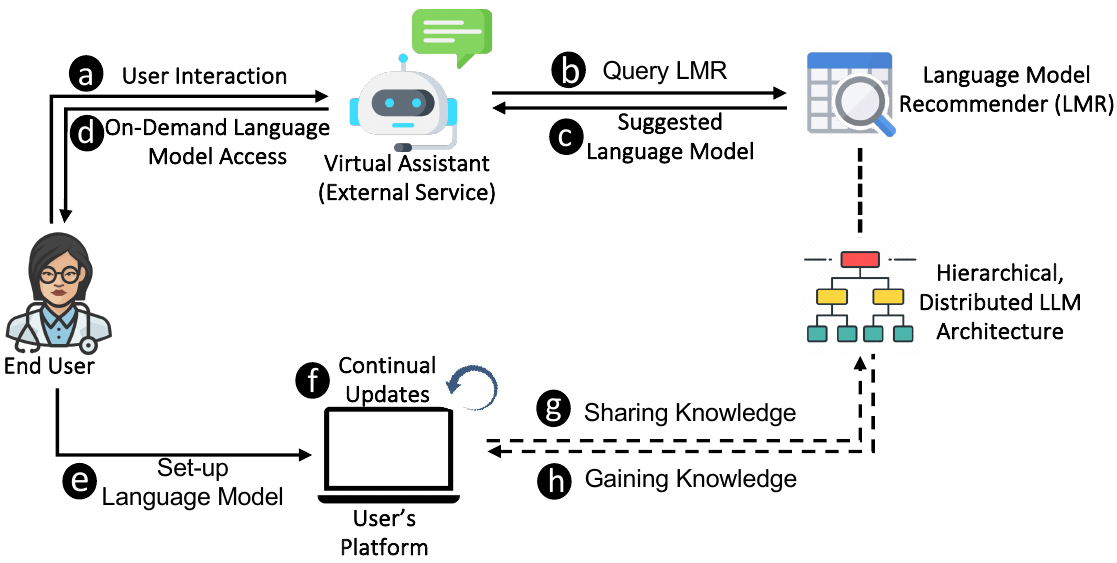}
        \caption{A Use Case - Leveraging Hierarchical Language Model Architecture as On-Demand Service.}
        \label{fig:use_case}
        \vspace{-3mm}
\end{figure*}

A  \textit{layered} architecture such as those proposed here
can offer seamless on-demand adoption of LLMs customized to a user's specific needs and available computing resources. Such a multi-layer architecture allows users more control when customizing LLMs according to their preferred platforms, languages, domains, and service preferences while ensuring the trade-off between the computational resources available and the business application needs. Furthermore,  the open-source implementation of this architecture plays a pivotal role in democratizing AI. Making it open and crowd-sourced promotes collaboration, accessibility, and a sense of shared ownership within the community. 
\section{Leveraging Adaptive Language Models for On-demand Service: A Healthcare Use Case}

Figure~\ref{fig:use_case} depicts a practical healthcare use case that exemplifies our conceptual architecture in resource-constrained environments.\footnote{Although we present a healthcare case study, our concept is general and applicable to various application domains.} 

\vspace{0.3\baselineskip}
\noindent 
\textbf{Scenario.}\quad Dr. Smith, a medical researcher at a rural healthcare facility equipped with devices of limited computational capabilities, is researching rare diseases. She needs advanced language models to analyze and categorize vast volumes of medical texts, research papers, and patient records. However, her local infrastructure lacks the computational power required to deploy and fine-tune LLMs, and she does not have sufficient funding to use a paid LLM service available online.

\vspace{0.3\baselineskip}
\noindent 
\textbf{Solution.}\quad Our hierarchical LLM architecture can aid Dr. Smith's study as follows.

\begin{enumerate}[label=\Circled{\alph*}, leftmargin=*, wide = 0pt]
  \item \textbf{User Interaction:} Dr. Smith initiates a conversation with a \textit{Virtual Assistant}, specifying her LLM service requirements and resource constraints.
  
  \item \textbf{Query Model Recommender:} The Virtual Assistant --- an ``online service agent'' --- notifies the \textit{Language Model Recommender} about Smith's specific requirements.
  
  \item \textbf{Recommendation:} The Language Model Recommender then finds the most suitable language model for Dr. Smith's research. The Recommender considers factors such as domain (healthcare), sub-domain (rare diseases), resource limitations, computational cost (processing time, memory), and monetary costs (to receive the service) associated with the language models.

  \item \textbf{On-Demand Access:} Dr. Smith, following the recommendation, acquires a suggested language model designed to match her hardware capacity and research requirements. This language model serves as an \textit{on-demand}, customizable research tool. We note that the chosen model is not rigid but highly \textit{adaptable}. For instance, Dr. Smith might initially acquire a general-purpose language model suitable for various medical topics. However, as her focus is on rare diseases requiring specific knowledge, she adapts the language model using her personal data corpus. This adaptability guarantees that the language model aligns with her research objectives, allowing her to delve into the intricacies of rare disease studies efficiently.
  
  \item \textbf{Local Language Model Setup:} Dr. Smith then deploys the language model on her local device (\eg a personal computer). 
  
  \item \textbf{Learning and Staying Updated:} Dr. Smith can update her language model \textit{continually}. For instance, let us assume Dr. Smith starts with a language model that knows up to a certain date, which is common in many commercial LLMs. Over time, she collects new research findings and data related to rare diseases. The language model can incorporate this new information into its knowledge base. For example, if a breakthrough discovery about a particular rare disease occurs, Dr. Smith can feed this information into the language model. The language model then learns from it, making it aware of the latest developments in the field. This ensures that Dr. Smith's research tool is always up-to-date and equipped with the most recent information, allowing her to stay at the forefront of her research field. 
  
  \item\textbf{Sharing Knowledge:} Whenever Dr. Smith updates her language model, the peer language models (part of the layered architecture) are notified. The peer language models ensure that the updates are efficiently shared with ``upstream'' and ``downstream'' language models. 
  
  \item\textbf{Gaining Knowledge:} When other peer language models update themselves, they share their knowledge with their counterparts. If the new information aligns with Dr. Smith's language model requirements, her model can efficiently gain this newfound knowledge from others.
\end{enumerate}


\noindent 
\textbf{Impact.}\quad By utilizing the hierarchical language model architecture, Dr. Smith can conduct advanced healthcare research even with \textit{limited computational resources}. The adaptability of language models to user-specific needs and the continual update process enhances user tasks (for instance, research capabilities of Dr. Smith), allowing the user to contribute towards a desired goal (\viz contribute to understanding and treating rare diseases in this case). The use case illustrates how our architecture democratizes access to advanced AI technology, reducing technological disparities.
\begin{figure*}[!tb]
        \centering
        \vspace{-1\baselineskip}
        \includegraphics[width=0.90\linewidth]{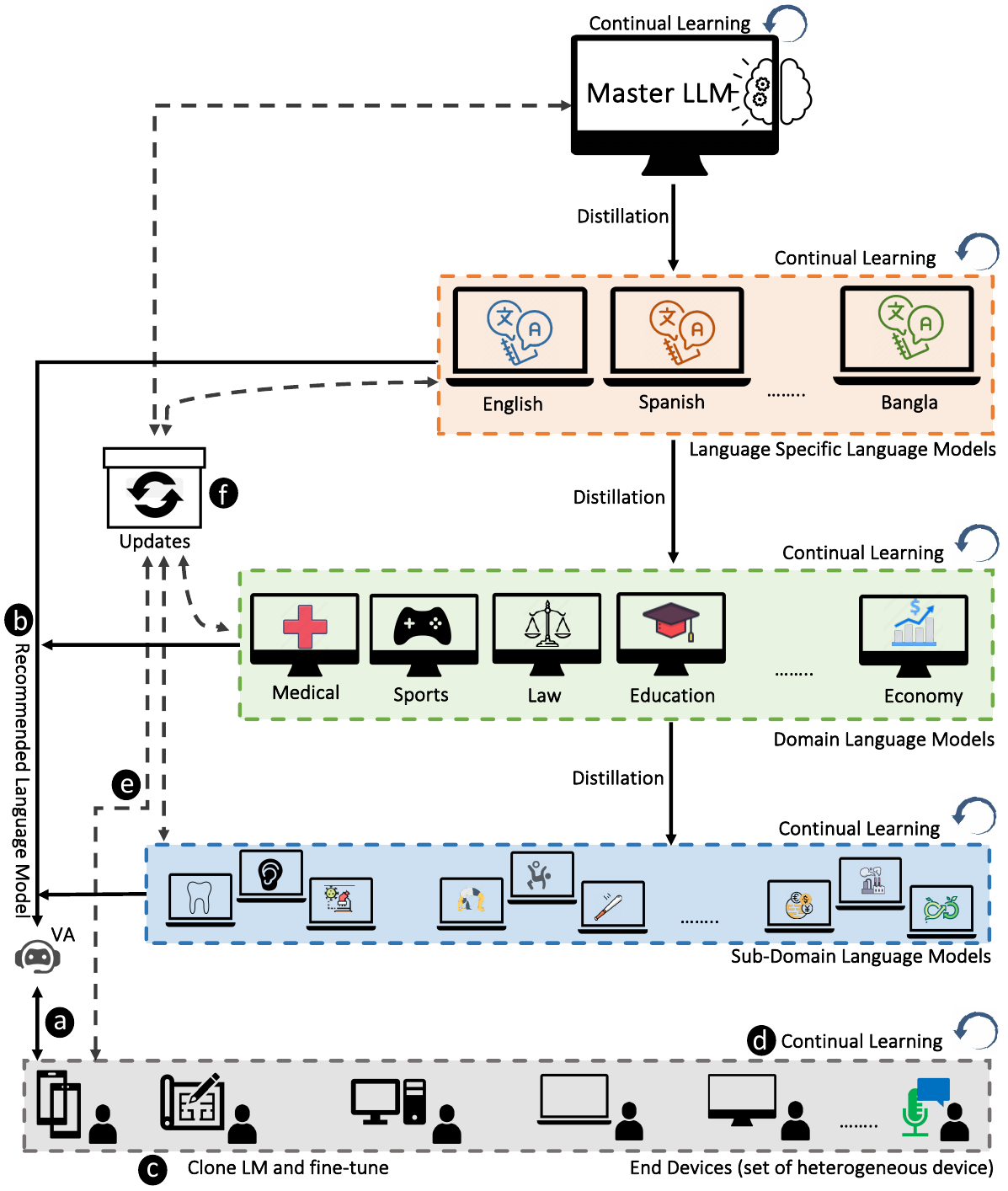}
        \caption{High-level schematic diagram of a multi-tier distributed LLM architecture.}
        \label{fig:dialog_system}
        \vspace{-3mm}
\end{figure*}

\section{A Multi-Layer LLM Architecture}\label{sec:architecture}


Figure~\ref{fig:dialog_system} presents an overview of the layered architecture. We propose to organize multiple language models in a \textit{hierarchical} order, considering languages, domains, sub-domains, variations in size, resource requirements, and computational cost. We arrange the language models in a ``top-down'' manner, with larger models at the top and smaller models at the bottom, following a decreasing order of size, resource availability, and computation cost. Such a hierarchical arrangement allows users to choose a language model that suits their needs and available resources. 

\subsection{Workflow}
The high-level workflow of our architecture is as follows.

    \textbf{i.~} The \textit{User} interacts with a \textit{Virtual Assistant}, providing specific requirements for their desired application.
    
    \textbf{ii.~} Based on the user's specifications, the virtual assistant consults a \textit{Language model Recommender System}~\cite{sarkar2023exploring} and recommends the most suitable language model for the scenario. Users with sufficient resources can opt for a larger language model, while those with limited computing power may choose smaller language models to achieve their goals.
    
    \textbf{iii.~} The user then \textit{clones} the recommended language model and proceeds to \textit{fine-tune} it on their goal task using their local devices. Fine-tuning means making small adjustments to an existing model to make it work better for a specific task or situation. For instance, consider a camera that can capture photos, but they do not always turn out perfect. Fine-tuning involves making small changes to the camera's settings, such as adjusting the focus or brightness, so that the photos appear just as the user wants. 
    
    \textbf{iv.~} The user can update the cloned language model with new data through \textit{continual learning}~\cite{sun2019lamol}. Continual learning (a.k.a. lifelong learning) enables AI systems to keep learning new things while retaining previous knowledge. This is akin to adding new information to an existing database without erasing existing information. For instance, when introducing a new animal, like a parrot, to an AI system trained to recognize cats and dogs, the system learns about parrots while still recognizing cats and dogs. This concept allows AI to build on its existing knowledge, making it smarter over time without forgetting what it has learned in the past.
    \vspace{-1mm}
    \textbf{v.~} When a language model is updated or fine-tuned with new data, the \textit{peer language models} are notified to update themselves with this new knowledge.
    
    \textbf{vi.~}
    The language model then shares this new knowledge with both the preceding and succeeding layer language models through \textit{Upstream} and \textit{Downstream} knowledge transfer mechanisms. Upstream knowledge transfer works as follows: imagine a group of researchers specializing in various subdomains of Biology, such as Microbiology and Zoology. They share their latest discoveries and findings with a senior researcher who acts as a mentor. This senior researcher works in a broader field, encompassing the specialized topics within Biology. In this context, knowledge flows from the junior researchers in subdomains like Microbiology and Zoology to the senior researcher, representing Upstream knowledge transfer. Now, let us delve into Downstream knowledge transfer. 
    Expanding on the scenario we discussed above, consider the role of a senior researcher in Biology. When the senior researcher comes across new and relevant information within specific subdomains of Biology, they take on the role of a knowledge distributor. In this context, knowledge flows from the senior to junior researchers specializing in particular areas of Biology, such as Microbiology and Zoology. 

\subsection{Components and Functions}

We now present the components and describe the functionalities of each layer.


\acmbfpar{User.} The user represents the end user who desires to obtain language model services based on their specific requirements and preferences.

\acmbfpar{Virtual Assistant (VA).} The VA interfaces the user and the backend layered architecture. The user interacts with the VA and provides specifications, such as the desired platform and services they are looking for. The VA then recommends the most suitable language model instance.

\acmbfpar{Master LLM Layer (Root).}  At the root of our hierarchical architecture resides the \textit{Master LLM}, the largest general-purpose language model available, and serves as the base (\viz \textit{``Teacher''}) model for transferring knowledge to successor language models. 


\acmbfpar{Language Specific Language Model (LSLM) Layer.} The following layer in the hierarchy is language models specific to a particular language (called LSLM, depicted by the Orange box in Fig.~\ref{fig:dialog_system}). LSLMs are smaller than the Master LLM. We can use distillation techniques~\cite{gou2021knowledge} across the hierarchy to transfer knowledge from a larger model to a smaller model. As an example, the Master LLM acts as the \textit{``Teacher''} and the language-specific (\eg English, Spanish) models as the \textit{``Student''} during the distillation process. 

\acmbfpar{Domain Language Model (DLM) Layer.} The subsequent layer contains domain-specific language models for each Language Specific Language Model (LSLM), such as Medical, Sports, Law, and Education, as shown by the Green box in Fig.~\ref{fig:dialog_system}). These domain-specific language models are \ci compact in size, \cii possess fewer parameters, and \ciii exhibit lower complexity. However, they possess the essential knowledge/information to excel in their respective domains in a specific language. 


\acmbfpar{Sub-Domain Language Model (SDLM) Layer.} The next layer of the architecture consists of SDLMs, essentially specialized language models tailored to specific sub-areas of a domain. For example, in the medical field, sub-domains include Virology or Heart Health. Likewise, in the sports industry, sub-domains may contain language models related to Gymnastics or Soccer. SDLMs can be customized to cater to the specific requirements of each domain precisely while ensuring optimal performance and usability. As we descend the hierarchy, these specialized models (illustrated by the Blue rectangle in Fig.~\ref{fig:dialog_system}) become increasingly focused, compact, and application-friendly.


\acmbfpar{End Devices Layer.} End devices include heterogeneous computing systems such as laptops, tablets, smartwatches, and embedded devices (as shown inside the Gray rectangle in Fig.~\ref{fig:dialog_system}). Depending on the specific application scenarios and requirements/constraints, a user can \ca acquire a preferred language model compatible with computing resources and \cb fine-tune it on their system depending on the goal task. 


\acmbfpar{Continual Learning.} Continual learning
is a machine learning paradigm where a model learns from a continuous stream of data over time. Unlike traditional machine learning, where models are typically trained on a static dataset and then tested on new data, continual learning models are designed to adapt and improve their performance as they encounter new data. In our setup, continual learning plays a pivotal role in updating the knowledge of the entire architecture. This allows the model to \ca adapt to new data and language patterns, \cb learn from recent examples, and \cc improve its performance over time, as discussed in our case study (Step \Circled{f}: Learning and Staying Updated). The open-source nature of the architecture plays a vital part in this process, as it encourages a collaborative effort among a diverse community. With a crowd-sourced community-driven approach, the model can adapt to the latest developments and benefit from a wealth of recent examples derived from various sources, including niche domains (based on user consent). This collective intelligence promotes constant improvements, leading to an architecture that continually refines its performance and consistently delivers more accurate and relevant results over time. As Fig.~\ref{fig:dialog_system} depicts, we leverage continual learning techniques~\cite{sun2019lamol} throughout the layer hierarchy (indicated by the circular arrow at the right corner of each layer). 



\acmbfpar{Upstream \& Downstream Knowledge Transfer.}
The bidirectional arrow in Fig.~\ref{fig:dialog_system} signifies the architecture's dynamic flow of updates and information exchange. When a language model engages in continual learning and updates itself, it triggers a two-way knowledge transfer process. This transfer occurs both Upstream (from bottom to top) and Downstream (from top to bottom) across all layers of language models. This collaborative exchange ensures that all models remain synchronized and can capitalize on the latest advancements and data insights. Although it is conceivable to perform knowledge transfer among language models within the same layer, this falls outside the current scope of our architecture.

\section{Challenges and Deployment Issues}\label{sec:challenges}

While our proposed framework provides one way to make LLMs customizable for heterogeneous devices, there exist several open deployment challenges, as we present below.



\acmbfpar{Challenge 1: How to identify the most suitable language model?}  
A comprehensive study of LLMs across diverse resource constraints and accuracy prerequisites is crucial. This endeavor enables users to make informed choices when opting for LLMs for applications, thus streamlining development efforts. Dr. Smith's scenario vividly underscores this challenge as she strives to employ advanced language models in medical research, even within her facility's resource limitations. This instance emphasizes the critical necessity for methodical language model assessments under varying resource constraints and accuracy criteria. Through this investigation, we aim to offer practical insights to assist users like Dr. Smith in optimizing their language model choices for local deployment. A comprehensive investigation, as conducted by ~\cite{karmaker2021automl}, can elucidate the pipeline of machine learning tasks and pinpoint the stages most vulnerable to resource constraints.

\acmbfpar{Challenge 2: How to coordinate continuous updates?} Effective collaboration among layers is vital to ensure a seamless process of continual learning. As depicted in Fig.~\ref{fig:dialog_system}, knowledge transfer is a dynamic process occurring in both \textit{Upstream} and \textit{Downstream}. To illustrate, when Dr. Smith updates her language model with new information, the language models in the hierarchy might receive notifications through upstream knowledge transfer based on some criteria. While downstream knowledge transfer can be achieved through distillation, achieving upstream knowledge transfer can be realized through three distinct methods: \ca sharing raw data, \cb employing a generator with synthetic training samples, or \cc sharing model parameters.

\noindent \textit{(a) \underline{Collaboration via Direct Dataset Sharing}:} When an end device retrains/updates its language model, the device can send the training dataset to the corresponding parent language model to update itself. However, this method raises concerns about data privacy since the original dataset is shared among language models.


\noindent \textit{(b) \ul{Privacy-Preserving Collaboration via Synthetic Data Sharing}}: To address the privacy concerns, a continual learning process with two components can be leveraged: the \textit{generator} and the \textit{learner}~\cite{sun2019lamol}. As bulk, new data arrives at the end devices, the learner component incorporates these generated examples into the training process, fine-tuning the model's parameters based on the new data. Next, The generator component generates synthetic training examples that mimic real-world language patterns and structures. These examples serve as additional training data for the peer language model, allowing it to learn from more diverse language variations and scenarios while preserving privacy. In cases where an intermediate language model (LSLM, DLM, SDLM) updates itself, its generator generates synthetic training examples and sends the dataset to the predecessor or successor layer for further updating. This approach ensures privacy and facilitates consistent, collaborative learning. In the healthcare research scenario, as new patient data arrives at healthcare facilities, the ``learner'' component incorporates it into the research process for fine-tuning/continual learning methods. Simultaneously, the ``generator'' component generates synthetic patient data, mimicking real-world patterns, and shares this with the research network. This approach ensures privacy and enables collaborative learning among healthcare researchers while maintaining data privacy.



\noindent \textit{(c) \ul{Privacy-Preserving Collaboration via Model} \underline{Parameter Sharing}:} Another alternative approach is that the parent language model maintains a copy of the immediate child (distilled) language models. Instead of sharing datasets, the updated model parameters are communicated back to the Upstream language model. This parameter-passing approach promotes collaborative learning while preserving data privacy, following the principles of federated learning~\cite{textfl,largelang}. For \textit{Downstream knowledge transfer}, a distilled \textit{Topic-Driven Generator} to generate topic-specific synthetic training data can be employed, which will be used to update the successor layer language models. However, there is a need to investigate specific techniques for Upstream and Downstream knowledge transfer to optimize the overall architecture.

\acmbfpar{Challenge 3: How to prevent the loss of previously learned knowledge?} One of the challenges we might face in this context is  the loss of previously learned knowledge. This issue is commonly referred to as ``catastrophic forgetting''~\cite{li2019learn}, where knowledge that was previously acquired is at risk of being forgotten when new information is introduced. Catastrophic forgetting is a common side-effect of continual learning, where previously acquired knowledge is at risk of being lost during the assimilation of new information. In the healthcare use case, catastrophic forgetting is pertinent if she continuously updates her language model with new medical data to stay current with rare disease research. The risk of forgetting critical insights while adapting to the latest information is a significant concern and needs further research in the context of hierarchical LLM architecture such as those we proposed here.

\vspace{-1mm}

\acmbfpar{Challenge 4: When should we update the parent language model?}  
One of the challenges 
is to define the criteria for updating an Upstream (parent) language model. Considering the multitude of end devices actively utilizing and modifying these models, managing this constant stream of updates is of utmost importance. 
End users may use specific language models, tailoring them to their immediate needs. However, not all updates are pertinent to the Master language model. Therefore, it becomes crucial to determine the optimal timing for these updates to maintain efficiency and effectiveness in the distributed architecture. One strategy could be evaluating the significance of new data in contributing to the broader knowledge base~\cite{ke2023continual}. 


\acmbfpar{Challenge 5: What if a node is malicious?} Thus far, we assume all nodes cooperate and obey the protocol. However, an end device could misbehave and try to destabilize the system by \ci data poisoning attacks (participants intentionally submit data with incorrect labels)~\cite{datapoison}, \cii model poisoning attacks (malicious participants modify the updated model directly before sending it for aggregation)~\cite{modelpoisoning}, and  \ciii free-riding attacks (malicious participants contribute passively)~\cite{freeriders}. These concerns are not just theoretical. A medical researcher, for instance, faces similar risks due to the dependency on labeled data for fine-tuning/continually updating models. If participants intentionally provide incorrect medical data or tamper with the model during updates, it could negatively affect research findings. Additionally, passive disruptions by malicious individuals could hinder the research process without their active participation. Hence, it is imperative to develop suitable techniques that can \ca isolate malicious nodes, \cb limit their information propagation to prevent poisoning Upstream models, and \cc identify if any misinformation colludes with parent models. 
 



\section{Conclusion}


This work presents a ``layered'' LLM architecture to tackle the challenges of deploying LLMs in practical, real-world applications. 
We believe this concept can serve as a stepping stone for implementing an open-source, customizable LLM architecture, which will foster a wider adoption of LLMs across various platforms and applications, unlocking their potential for agile performance. 

\def\refname{REFERENCES}

\bibliographystyle{IEEEtran}
\bibliography{main}

\end{document}